# Semantic Channel and Shannon's Channel Mutually Match for Multi-Label Classification

Chenguang Lu [0000-0002-8669-0094]

`lcguang@foxmail.com`

**Abstract.** A group of transition probability functions form a Shannon's channel whereas a group of truth functions form a semantic channel. Label learning is to let semantic channels match Shannon's channels and label selection is to let Shannon's channels match semantic channels. The Channel Matching (CM) algorithm is provided for multi-label classification. This algorithm adheres to maximum semantic information criterion which is compatible with maximum likelihood criterion and regularized least squares criterion. If samples are very large, we can directly convert Shannon's channels into semantic channels by the third kind of Bayes' theorem; otherwise, we can train truth functions with parameters by sampling distributions. A label may be a Boolean function of some atomic labels. For simplifying learning, we may only obtain the truth functions of some atomic label. For a given label, instances are divided into three kinds (positive, negative, and unclear) instead of two kinds as in popular studies so that the problem with binary relevance is avoided. For each instance, the classifier selects a compound label with most semantic information or richest connotation. As a predictive model, the semantic channel does not change with the prior probability distribution (source) of instances. It still works when the source is changed. The classifier changes with the source, and hence can overcome class-imbalance problem. It is shown that the old population's increasing will change the classifier for label "Old" and has been impelling the semantic evolution of "Old". The CM iteration algorithm for unseen instance classification is introduced.

## 1 Introduction

There have been many valuable studies [1-6] about multi-label classification. Information, cross-entropy, and uncertainty criterions also have been used [7-9]. This study inherits some methods presented by others. However, this study aims at: 1) Letting machine learning from big enough samples for the semantic meanings (truth functions) of labels; 2) Decomposing the multi-label learning task into a number of independent label learning tasks without problem with binary relevance; 3) Overcoming the class imbalance problem when the prior probability distribution of instances is variable; 3) Adhering to Maximum Likelihood (ML) criterion by using Maximum Semantic Information (MSI) criterion that is compatible with ML criterion and close to Regularized Least Squares criterion.

In poplar machine learning studies, Bayesian inference is often used. However, this study uses a pair of new Bayes' formulas (i. e. Bayes' Theorem III) for setting up the conversion relation between likelihood functions and truth functions, and uses truth functions to describe membership relations between instances and classes.

In recent two decades, the cross-entropy method has become popular [4]. However, Lu [10] proposed the cross mutual information defined with the cross entropy as early as 199. He also proposed a matching function $R(G)$ between Shannon's mutual information and semantic mutual information or average log-normalized-likelihood [11-12] for the optimization of semantic communication. Recently, we found this function was very useful to semi-supervised learning [13] and unsupervised leaning [14]. We also found the new method was useful to multi-label classification.

The rest of this paper is organized as follows. The next section provides mathematical methods including the third kind of Bayes' theorem and the semantic information method. Section 3 discusses how new methods are applied to multi-label classifications for clear classes and fuzzy classes and how the classifier with the semantic information criterion changes with the prior probability distribution of instances for overcoming the class-imbalance problem. Section 4 simply introduces the CM iteration algorithm for semi-supervised classification with unseen instences. Section 5 includes some discussions and conclusions.

# 2 Mathematical Methods

## 2.1 Distinguishing Statistical Probability and Logical Probability

**Definition 2.1.1** Let $U$ denote the instance set and $X$ denote the discrete random variable taking a value from $U$. That means $X \in U = \{x_1, x_2, …\}$. For theoretical convenience, we assume that $U$ is one-dimensional. Let $L$ denote the set of selectable labels, including some atomic labels and some compound labels and $Y \in L = \{y_1, y_2, …\}$. Similarly, let $L_a$ denote the set of some atomic labels and $a \in L_a = \{a_1, a_2, …\}$.

**Definition 2.1.2** A label $y_j$ is also as a predicate $y_j(X) = $ "$X \in A_j$". For each $y_j$, $U$ has a subset $A_j$, every instance of which makes $y_j$ true. Let $P(Y=y_j)$ denote the statistical probability of $y_j$, and $P(X \in A_j)$ denote the Logical Probability (LP) of $y_j$. For simplicity, let $P(y_j) = P(Y=y_j)$ and $T(y_j) = T(A_j) = P(X \in A_j)$.

We call $P(X \in A_j)$ the logical probability because according to Tarski's theory of truth [15], $P(X \in A_j) = P($"$X \in A_j$" is true$) = P(y_j$ is true$)$. Hence the conditional LP of $y_j$ for given $X$ is the feature function of $A_j$ and the truth function of $y_j$. Let it be denoted by $T(A_j|X)$. There is

$$T(A_j) = \sum_i P(x_i) T(A_j | x_i) \tag{2.1}$$

According Davidson's truth-conditional semantics [16], $T(A_j|X)$ ascertains the semantic meaning of $y_j$. Note that statistical probability distribution, such as $P(Y)$, $P(Y|x_i)$, $P(X)$, and $P(X|y_j)$, are normalized whereas the LP distribution is not normalized. For example, in general,

$$T(A_1) + T(A_2) + … + T(A_n) \geqslant 1, \ T(A_1|x_i) + T(A_2|x_i) + … + T(A_n|x_i) \geqslant 1$$

$P(A_j) = T(A_j)$ only when $\{A_1, A_2, …, A_n\}$ is a partition of $U$ and $Y$ is always correctly used. $T(A_j|X)$ is similar to $P(y_j|X)$; yet its maximum is 1.

If $A_j$ may be fuzzy [17], let $A_j$ be replaced with $\theta_j$, which means a fuzzy set or fuzzy class. The $\theta_j$ can also be regarded as a sub-model of a predictive model $\theta$. In this paper, likelihood function $P(X|\theta_j)$ is equal to $P(X|\theta, y_j)$ in popular likelihood method. The usage of the sub-model $\theta_j$ will make the predictive model flexile and the statements clearer.

## 2.2 Three Kinds of Bayes' Theorems

The Bayes' theorem is described by Bayes' formulas. Actually, this theorem has three forms, which are used by Bayes [18], Shannon [19], and the author [12] respectively.

**Bayes' Theorem I** (used by Bayes): Assume that sets $A, B \in 2^U$, $A^c$ is the complementary set of $A$, $T(A) = P(X \in A)$, and $T(B) = T(A) = P(X \in B)$. Then

$$T(B|A) = T(A|B)T(B)/T(A), \ T(A) = T(A|B)T(B) + T(A|B^c)T(B^c) \tag{2.2}$$

$$T(A|B) = T(B|A)T(A)/T(B), \ T(B) = T(B|A)T(A) + T(B|A^c)T(A^c) \tag{2.3}$$

Note there is only one random variable $X$ and two logical probabilities.

**Bayes' Theorem II** (used by Shannon): Assume that $X \in U$, $Y \in V$, $P(x_i) = P(X=x_i)$, and $P(y_j) = P(Y=y_j)$. Then

$$P(x_i | y_j) = P(y_j | x_i) P(x_i) / P(y_j), \ P(y_j) = \sum_i P(x_i) P(y_j | x_i) \tag{2.4}$$

$$P(y_j | x_i) = P(x_i | y_j) P(y_j) / P(x_i), \ P(x_i) = \sum_j P(y_j) P(x_j | y_j) \tag{2.5}$$

Note there are two random variables and two statistical probabilities. In each of the above two theorems, two formulas are symmetrical and denominators are normalizing constants.

**Bayes' Theorem III:** Assume that $P(X)=P(X=\text{any})$ and $T(A_j)=P(X\in A_j)$. Then

$$P(X|A_j) = T(A_j|X)P(X)/T(A_j), \quad T(A_j) = \sum_i P(x_i)T(A_j|x_i) \quad (2.6)$$

$$T(A_j|X) = P(X|A_j)T(A_j)/P(X), \quad T(A_j) = 1/\max(P(X|A_j)/P(X)) \quad (2.7)$$

The two formulas are asymmetrical because there is a statistical probability and a logical probability. $T(A_j)$ in (2.7) may be call longitudinally normalizing constant. The proof is provided in another paper [20]. When the set $A_j$ becomes a fuzzy set $\theta_j$, the Bayes' theorem III is still tenable.

## 2.3  Distinguishing Statistical Probability and Logical Probability

In Shannon's information theory [19], $P(X)$ is called the source and $P(Y)$ is called the destination, the transition probability matrix $P(Y|X)$ is called the channel. So, a channel is formed by a group of transition probability function:

$$P(Y|X): P(y_j|X), j=1, 2, …, n$$

Note that $P(y_j|X)$ is different from $P(Y|x_i)$; $P(y_j|X)$ ($y_j$ is constant and $X$ is variable) is also not normalized. It has two important properties: 1) $P(y_j|X)$ can be used for Bayes' prediction to get $P(X|y_j)$; after $P(X)$ becomes $P'(X)$, $P(y_j|X)$ still works for the prediction; 2) $P(y_j|X)$ by a constant $k$ can make the same prediction because

$$\frac{P'(X)kP(y_j|X)}{\sum_i P'(x_i)kP(y_j|x_i)} = \frac{P'(X)P(y_j|X)}{\sum_i P'(x_i)P(y_j|x_i)} = P'(X|y_j) \quad (2.8)$$

Similarly, a group of truth functions form a semantic channel:

$$T(\theta|X): T(\theta_j|X), j=1, 2, …, n$$

According to (2.8), if $T(\theta_j|X) \propto P(y_j|X)$ or $T(\theta_j|X) = P(y_j|X)/\max(P(y_j|X))$, there is $P(X|\theta_j) = P(X|y_j)$.

## 2.4  To Define semantic information with log (normalized likelihood)

The (amount of) semantic information conveyed by $y_j$ about $x_i$ is defined with log-normalized-likelihood [10, 11]:

$$I(x_i;\theta_j) = \log \frac{P(x_i|\theta_j)}{P(x_i)} = \log \frac{T(\theta_j|x_i)}{T(\theta_j)} \quad (2.9)$$

where Bayes' Theorem III is used. For an unbiased estimation $y_j$, its truth function may be assumed to be a Gaussian distribution without coefficient:

$$T(\theta_j|X) = \exp[-(X-x_j)^2/(2d_2)] \quad (2.10)$$

Then $I(x_i; \theta_j)$ changes with $x_i$ as shown in Fig. 1. This information criterion reflects Popper's thought [21]. It tells that the larger the deviation is, the less information there is; the less the logical probability is, the more information there is; and, a wrong estimation may convey negative information.

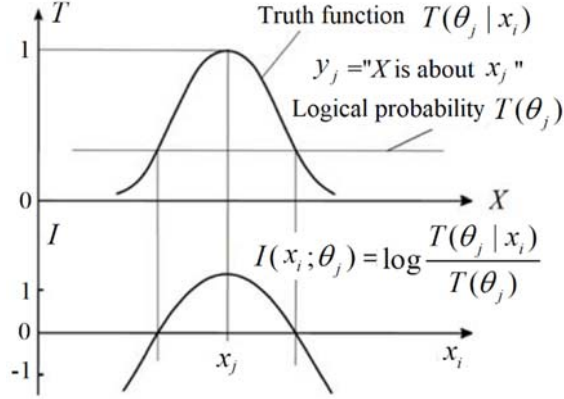

**Fig. 1.** Illustration of semantic information measure.

To average $I(x_i; \theta_j)$, we have

$$I(X;\theta_j) = \sum_i P(x_i|y_j)\log\frac{P(x_i|\theta_j)}{P(x_i)} = \sum_i P(x_i|y_j)\log\frac{T(\theta_j|x_i)}{T(\theta_j)} \quad (2.11)$$

$$I(X;\theta) = \sum_j P(y_j)\sum_i P(x_i|y_j)\log\frac{P(x_i|\theta_j)}{P(x_i)}$$
$$= \sum_j\sum_i P(x_i, y_j)\log\frac{T(\theta_j|x_i)}{T(\theta_j)} = H(\theta) - H(\theta|X) \quad (2.12)$$
$$H(\theta) = -\sum_j P(y_j)\log T(\theta_j),\; H(\theta|X) = -\sum_j\sum_i P(x_i, y_j)\log T(\theta_j|x_i)$$

where $I(X; \theta_j)$ is generalized Kullback-Leibler (KL) information, and $I(X; \theta)$ is the semantic mutual information (a mutual cross-entropy). It is easy to find that when $P(x_i|\theta_j)=P(x_i|y_j)$ for all $i, j$, $I(X; \theta)$ reaches its upper limit: Shannon mutual information $I(X; Y)$. To bring (2.10) into (2.12), we have

$$I(X;\theta) = H(\theta) - H(\theta|X)$$
$$= -\sum_j P(y_j)\log T(\theta_j) - \sum_j\sum_i P(x_i, y_j)(x_i - x_j)^2/(2d_j^2) \quad (2.13)$$

It is easy to find that the maximum semantic mutual information criterion is similar to the regularized least squares criterion. $H(\theta|X)$ is simillar to mean squared error and $H(\theta)$ is simillar to negative regularization term.

Assume that a sample is $D=\{(x(t); y(t)|t=1, 2, …, N; x(t)\in U; y(t)\in V\}$, a conditional sample is $\{x(1), x(2), …, x(N_j)\}$ for given $y_j$, and the sample points come from independent and identically distributed random variables. If $N_j$ is big enough, then $P(x_i|y_j)= N_{ij}/N_j$ where $N_{ij}$ is the number of $x_i$ in $D$. Then we have the log normalized likelihood:

$$\log\prod_i\left[\frac{P(x_i|\theta_j)}{P(x_i)}\right]^{N_{ji}} = N_j\sum_i P(x_i|y_j)\log\frac{P(x_i|\theta_j)}{P(x_i)} = N_j I(X;\theta_j) \quad (2.14)$$

To average $I(X; \theta_j)$ for different $y_j$, we will have the formula: Average log(normalized likelihood)=Semantic mutual information. Since $P(X)$ is irrelevant to $\theta_j$, the maximum semantic information criterion is equivalent to the maximum likelihood criterion.

## 3  Multi-Label Classification for Seen Instances

### 3.1  Multi-Label Learning (the Receiver's Logical Classification) for Truth Functions without Parameters

From the viewpoint of semantic communication, the sender's classification and the receiver's classification are different. The receiver learns from samples to obtain labels' semantic meanings, i. e. the truth functions. The learning means logical classification. Then, when he receives $y_j$, he can predict $X$ to obtain $P(X|\theta_j)$ according to $P(X)$ and $T(\theta_j|X)$ so that he can make a decision. However, for given an instance, the sender needs to select a label with most information from several true or truer labels. This is selective classification. We may say that the logical classification is for the denotations of labels and selective classification is for connotations of labels; the learning is letting a semantic channel match a Shannon's channel whereas the selectin is letting a Shannon's channel match a semantic channel

We use an example to show the two kinds of classifications. Assume that $U$ is a set of different ages. There are subsets $A_1$={young people}={$X$|15≤$X$≤35}, $A_2$={adults}={$X$|$X$≥18}, $A_3$={juveniles}={$X$|$X$<18}=$A_2^c$ ($^c$ means complementary set) of $U$, which form a cover of $U$. Three truth functions $T(A_1|X)$, $T(A_2|X)$, and $T(A_3|X)$ represent the semantic meanings of $y_1$, $y_2$, and $y_3$ respectively, as shown in Fig. 2.

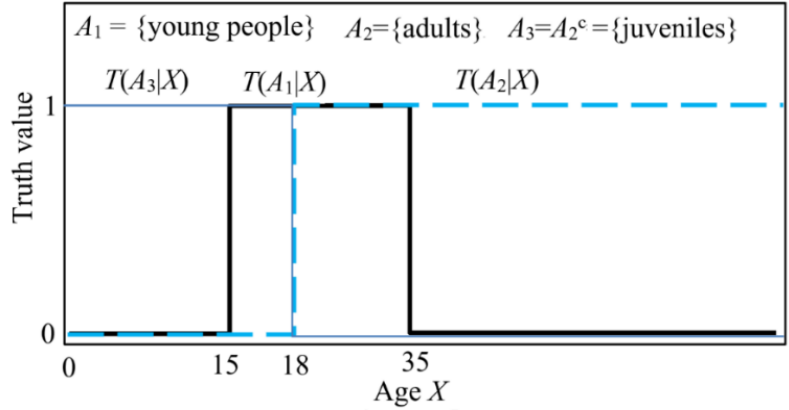

**Fig. 2.** Three sets form a cover of $U$, indicating the semantic meanings of $y_1$, $y_2$, and $y_3$.

In this example, $T(A_2)+T(A_3)=1$. If $T(A_1)=0.3$, then the sum of the three logical probabilities is 1.3>1. Yet, the sum of three statistical probabilities $P(y_1)+P(y_2)+P(y_3)$ must be 1. $P(y_1)$ may changes from 0 to 0.3. For example, $P(y_1)=0.2$, $P(y_2)=0.5$, and $P(y_3)=0.3$.

Using the truth functions $T(A_j|X)$ or $T(\theta_j|X)$ instead of Bayesian probability $P(\theta, y_j|X)$ or $P(\theta_j|X)$ for multi-label learning, we can independently obtain the truth function of a label without considering $\Sigma_j P(y_j|X)=1$ or $\Sigma_j P(\theta_j|X)=1$.

**Definition 2.3.1**  For a sampling $D$ with distribution $P(X)$, all instances in $A_j$ of $D$ form a window sample of $D$ with distribution $P(X|y_j)$, which should be equal to $P(X|A_j)$. We call the $P(X|y_j)$ a window distribution of $P(X)$.

**Theorem 2.3.1**  If $P(X)$ and $P(X|y_j)$ come from the same sample $D$, then $P(X|y_j)$ is a window distribution of $P(X)$. If $D$ is big enough so that every possible example appears at least one time, then we can directly obtain the numerical solution of feature function of $A_j$ (as shown in Fig. 3 (a)) according to Bayes' Theorem III and II:

$$T^*(A_j|X) = \frac{P(X|y_j)}{P(X)} \bigg/ \max\left(\frac{P(X|y_j)}{P(X)}\right) = P(y_j|X)/\max(P(y_j|X)) \qquad (3.1)$$

The proof is omitted. It is easy to prove that changing $P(X)$ and $P(Y)$ does not affect $T^*(A_j|X)$ because $T^*(A_j|X)$ ($j$=1, 2, …) reflect the property of Shannon's channel or the semantic channel. This formula is also tenable to fuzzy set $\theta_j$.

If $P(X|y_j)$ is from another sampling instead of the window sample of $P(X)$, then $T^*(A_j|X)$ will not be smooth as shown in Fig. 3 (b). The larger the size of $D$ is, the smoother the truth function is.

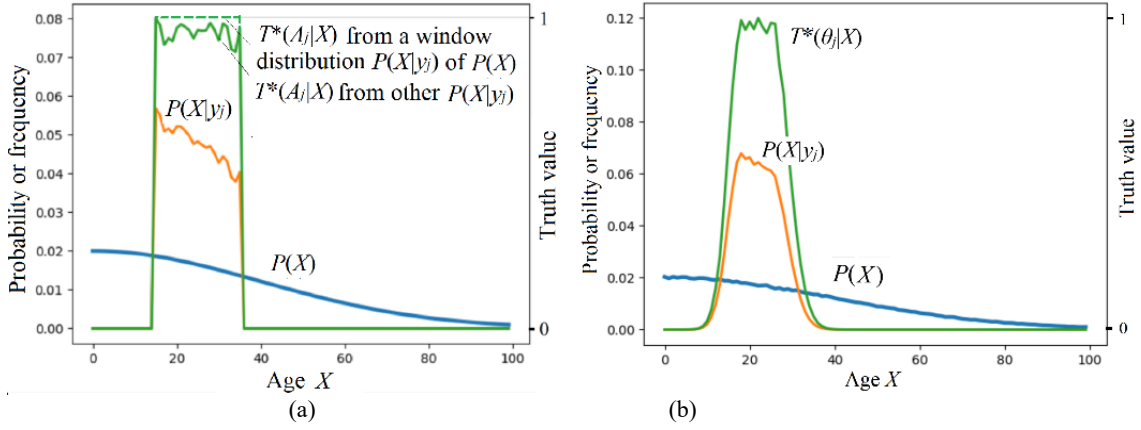

**Fig. 3.** The numerical solution of the membership function according to (3.1); (a) for a crisp set and (b) for a fuzzy set.

### 3.2 Selecting Examples for Atomic Labels

According to mathematical logic, $k$ atomic propositions may produce $2^k$ independent clauses. The logical add of some of them has $2^{2**k}$-2 results, neglecting contradiction and tautology. So, there are $2^{2**k}$-2 possible compound labels. To simplify the learning, we may filter examples in a multi-label sample to form a new sample $D_a$ with $k$ atomic labels and $k$ corresponding negative labels. We may use First-Order-Strategy [6-8] to split examples in $D$ with multi-labels or multi-instances into simple examples, such as, to split ($x_1$; $a_1$, $a_2$) into ($x_1$; $a_1$) and ($x_1$; $a_2$), and to split ($x_1$, $x_2$; $a_1$) into ($x_1$; $a_1$) and ($x_2$; $a_1$) [21]. Let $Y_a$ denote one of the $2k$ labels, i. e. $Y_a \in \{a_1, a_1', a_2, a_2', …, a_k, a_k'\}$. Consider some $a_j'$ does not appears in $D_a$, $Y_a$ may be one of $k+k'$ ($k'<k$) labels. From $D_a$, we can obtain $P(X, Y_a)$ and corresponding semantic channel $T^*(\theta_a|X)$ or $T^*(\theta_{aj}|X)$ ($j$=1,2, …, $2k$). If $D_a$ is big enough, we can obtain the numerical solution of $T(\theta_a|X)$ by (3.1); otherwise, we need to construct $T(\theta_a|X)$ by some parameters and to train it by $P(X|Y_a)$ and $P(X)$.

### 3.3 Multi-label Learning for Truth Functions with Parameters

If $P(Y, X)$ is obtained from a not big enough sample, we can optimize the truth function with parameters of every compound label by

$$T^*(\theta_j | X) = \arg\max_{T(\theta_j|X)} I(X;\theta_j) = \arg\max_{T(\theta_j|X)} \sum_i P(x_i | y_j) \log \frac{T(\theta_j | x_i)}{T(\theta_j)} \qquad (3.2)$$

It is easy to prove that when $P(X|\theta_j)=P(X|y_j)$, $I(X; \theta_j)$ reaches the maximum and is equal to the KL information $I(X; y_j)$. So, the above formula is compatible with (3.1). Comparing two truth functions, we can find logical implication between two labels. If $T(\theta_j|X) \leq T(\theta_k|X)$ for every $X$, then $y_j$ implies $y_k$, and $\theta_j$ is the subset of $\theta_k$. We may use logistic function, Gaussian function without coefficient, and other functions as truth functions with parameters. The detailed discussions will be provided later.

If we only want to obtain the truth functions of some atomic labels because $D$ is too big, we can optimize an atomic label by

$$T^*(\theta_{aj} | X) = \underset{T(\theta_j|X)}{\arg\max}[I(X;\theta_{aj}) + I(X;\theta_{aj}^c)]$$

$$= \underset{T(\theta_{aj}|X)}{\arg\max} \sum_i [P(x_i | a_j) \log \frac{T(\theta_{aj} | x_i)}{T(\theta_{aj})} + P(x_i | a_j') \log \frac{T(\theta_{aj}^c | x_i)}{T(\theta_{aj}^c)}] \quad (3.3)$$

where $T(\theta_{aj}^c|x_i)=1-T(\theta_{aj}|x_i)$. $T^*(\theta_{aj}|x_i)$ is only affected by $P(a_j|X)$ and $P(a_{j'}|X)$. Although those examples without $a_j$ or $a_{j'}$ affect $T^*(\theta_{aj})$, they do not affect $T^*(\theta_{aj}|x_i)$. For a given label, this method actually divides all instances into three kinds: the positive, the negative, and the unclear. $T^*(\theta_{aj}|x_i)$ is not affected by unclear instances.

If a negative label $a_j'$ does not appear in $D$ or $D_a$, the second part will be 0. So, this binary classification is different from popular One vs Rest [3] classification, with which the problem is that an example $(x_i, a_1)$ without label $a_2$ does not means that $x_i$ makes $a_2'$ true. For example, $x_i=25$, $x_i$ makes both $a_1$="youth" and $a_2$="adult" true.

We may also train $T(\theta_{aj}|x_i)$ and $T(\theta_{aj'}|x_i)$ separately. Then $T(\theta_{aj}|X)+T(\theta_{aj'}|X)=1$ will not be tenable. In many cases, we use three or more labels rather than two to tag one dimension of instance spaces, or each label is not a strictly negative label of another, the formula (3.2) is still suitable. For example, the truth functions of "Child", "Youth", and "Adult" may be separately optimized by three conditional sampling distributions. We needn't decompose a multi-class classification into several binary classifications. The following classifier $h(X)$ will resolve instance space partition problem.

## 3.4 Multi-label Selection (the Sender's Selective Classification)

For a seen instance $X$, the label sender selects $y_j^*$ by classifier

$$y_j^* = h(X) = \underset{y_j}{\arg\max} \log I(\theta_j; x_i) = \underset{y_j}{\arg\max} \log \frac{T(\theta_j | X)}{T(\theta_j)} \quad (3.4)$$

$T(\theta_j)$ is related to the class-imbalance. If $T(\theta_j|X) \in \{0, 1\}$, the information measure becomes Bar-Hillel and Carnap's information measure [22]; the classifier becomes

$$y_j^* = h(X) = \underset{y_j}{\arg\max}_{T(A_j|X)=1} \log[1/T(A_j)] = \underset{y_j}{\arg\min}_{T(A_j|X)=1} T(A_j) \quad (3.5)$$

For $X=x_i$, If several labels are correct or approximatively correct, we select one from $2^k$ independent clauses with maximum $I(x_i; \theta_j)$. For example, when $k=2$, these clauses are $a_1 \wedge a_2$, $a_1 \wedge a_2'$, $a_1' \wedge a_2$, and $a_1' \wedge a_2'$.

When sets are fuzzy, we may use a little different fuzzy logic [10] from Zadeh' [17] so that a compound label is a Boolean function of some atomic labels. There is

$$T(\theta_1 \cap \theta_2^c | X) = \max(0, \ T(\theta_1|X) - T(\theta_2|X)) \quad (3.6)$$

so that $T(\theta_1 \cap \theta_1^c|X)=0$ and $T(\theta_1 \cup \theta_1^c|X)=1$. Fig. 4 shows the truth functions of $2^2$ independent clauses, which form a partition of plan $U^*[0,1]$.

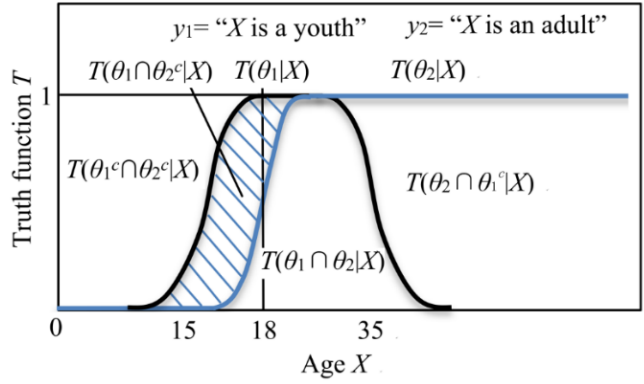

**Fig. 4.** The truth functions of $2^2$ independent clauses

### 3.5 How the Classifier $h(X)$ Changes with $P(X)$ to Overcome Class-imbalance Problem

Although optimized truth function $T^*(\theta_j|X)$ does not change with $P(X)$, the classifier $h(X)$ changes with $P(X)$. Assume that $y_4$="Old person", $T^*(\theta_4|X)=1/[1+\exp(-0.2(X-75))]$, $P(X)=1-1/[1+\exp(-0.15(X-c))]$. The $h(X)$ changes with $c$ as shown in Table 1.

**Table 1.** The classifier $h(X)$ for $y_4$="Old person" changes with $P(X)$

| $c$ | Population density decreasing ages | Classifier |
|---|---|---|
| 50 | 40-60 | $y_4=h(X>48)$ |
| 60 | 50-70 | $y_4=h(X>54)$ |
| 70 | 60-80 | $y_4=h(x>57)$ |

The dividing point of $h(X)$ increases when old population increases because semantic information criterion encourages us to reduce the failure of reporting small probability events. Longevous population's increasing changes $h(X)$; new $h(X)$ will change Shannon's channel and produce new samples; new semantic channel will match new Shannon's channel… The semantic meaning of "Old" should have been evolving with human lifetimes in this way. Meanwhile, the class-imbalance problem is overcome.

## 4 The CM Iteration Algorithm for Multi-Label Classification of Unseen Instances

For unseen instances, assume that observed condition is $Z \in C=\{z_1, z_2, …\}$; the classifier is $Y=f(Z)$; a true class or true label is $X_L \in U_L=\{X_1, X_2, …\}$; a sample is $Dz=\{(X(t); z(t)|t=1, 2, …, N; X(t) \in U_L; z(t) \in C\}$. From $Dz$, we can obtain $P(X_L, Z)$. If $Dz$ is not big enough, we may also use the likelihood method to obtain $P(X_L, Z)$ with parameters. The problem is that Shannon's channel is not fixed and also needs optimization. Hence, we treat unseen instance learning as semi-supersized learning. Using the channels' matching iteration algorithm or the CM iteration algorithm [13, 14], we can find optimal Shannon's channel and Semantic channel at the same time.

Let $C_j$ be a subset of $C$ and $y_j=f(Z|Z \in C_j)$. Hence $S=\{C_1, C_2, …\}$ is a partition of $C$. Our aim is, for given $P(X, Z)$ from $Dz$, to find optimal $S$, which is

$$S^* = \arg\max_S I(X;\theta|S) = \arg\max_S \sum_j \sum_i P(C_j)P(x_i|C_j)\log\frac{T(\theta_j|x_i)}{T(\theta_j)} \qquad (4.1)$$

First, we obtain the Shannon channel for given $S$:

$$P(y_j | X_i) = \sum_{z_k \in C_j} P(z_k | X_i), \ i=1,2,...,n; j=1,2,...,n \quad (4.2)$$

From this Shannon's channel, we can obtain the semantic channel $T(\theta|X_L)$ in numbers or with parameters. For given $Z$, we have conditional semantic information

$$I(X_i; \theta_j | Z) = \sum_i P(X_i | Z) \log \frac{T(\theta_j | X_i)}{T(\theta_j)} \quad (4.3)$$

Then let the Shannon channel match the semantic channel by

$$P(y_j | Z) = \lim_{s \to \infty} \frac{[\exp(I(X_L; \theta_j | Z))]^s}{\sum_{j'} [\exp(I(X_L; \theta_{j'} | Z))]^s}, \ j=1, 2, ..., n \quad (4.4)$$

Since $s \to \infty$, $P(y_j|Z)=0$ or 1. Hence, the above formula provides a classifier $Y=f(Z)$ and a new $S$. Repeating (4.2)-(4.4) until $S$ does not change. The convergent $S$ is the $S^*$ we seek. Some iterative examples show that the above algorithm is fast and reliable. The convergence can be proved with the help of the $R(G)$ function as shown in Fig. 5.

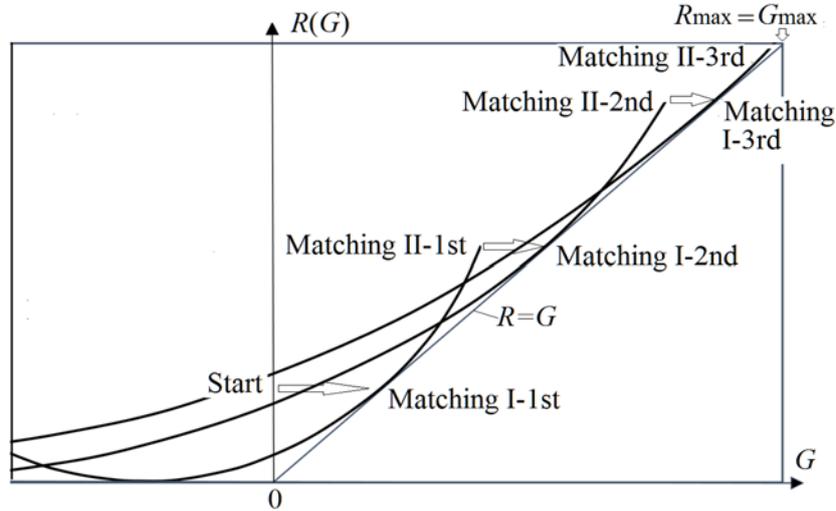

Fig. 5. Illustrating the iterative convergence for tests, estimations, and unseen instance classifications. The matching I is for $G=R$. The matching II is to increase $R$ to the top-right corner of a $R(G)$ function. Repeating the matching I and matching II can maximize $R$ and $G$ to obtain $R_{max}$ and $G_{max}$.

For more details, see [13].

## 5 Discussions and Conclusions

This paper brings truth functions of labels into new Bayes' formulas to produce likelihood functions. With the semantic information method (SIM), for very big samples, we can directly convert sampling distribution into truth functions for semantic meanings of labels. For not big enough samples, we can train truth functions with parameters by sampling distributions. When the prior distribution of instances is changed, the truth function as predictive model still works. Multi-label classification is distinguished into receivers' logical classification (learning) and senders' selective classification so that the logical classification is much simple. We discuss how the classifier changes with the prior distribution of in-stances and how class-imbalance problem is overcome. Unseen instance classification is treated as semi-supervised classification and can be resolved by the Channel Matching (CM) iteration algorithm.

The SIM is a challenge to Bayesian inference. The comparison between them needs further discussions. Regularized Least Squares criterion is getting popular. It seems that maximum likelihood criterion is out of date. However, this paper shows that the maximum semantic information criterion is compatible with the both. The mutual matching of semantic channel and Shannon's channel is very similar to the mutual contest of generator and discriminator in GAN [24]. The relationship between the two methods is worth analyzing. The SIM is also compatible with Wittegenstein's viewpoint that the meaning of a word lies its use [25]. In this paper, the instance space is assumed to be one dimensional. The optimization of truth functions with parameters for multidimensional instance space needs further studies.

The application of the CM iteration algorithm to mixture models or non-supervised learning also shows its advances [14] in comparison with the EM algorithm. It seems that the CM algorithm has wide potential applications. The SIM is concise and compatible with the thoughts of Shannon, Fisher, Popper, Wittegenstein, Bayes, Zadeh, Tarski, and Davidson. It should be a competitive alternative to Bayesian inference.